\documentclass[conference]{IEEEtran}
\usepackage{cite}
\usepackage{graphicx}
\usepackage{multirow}
\usepackage{dirtytalk}
\usepackage{hyperref}
\graphicspath{ {images/} }

\begin{document}


\title{Detecting Hate Speech and Offensive Language on Twitter using Machine Learning: An N-gram and TFIDF based Approach}


\author{
    \IEEEauthorblockN{
    Aditya Gaydhani\IEEEauthorrefmark{1},
    Vikrant Doma\IEEEauthorrefmark{2},
    Shrikant Kendre\IEEEauthorrefmark{3}
    and Laxmi Bhagwat\IEEEauthorrefmark{4}
    }
    \IEEEauthorblockA{Department of Computer Engineering, Maharashtra Institute of Technology, Pune\\
    Pune, India\\
    Email: \IEEEauthorrefmark{1}aditya.gaydhani@gmail.com,
    \IEEEauthorrefmark{2}vikrant.doma@gmail.com,
    \IEEEauthorrefmark{3}siddd.shriii@gmail.com,
    \IEEEauthorrefmark{4}laxmi.bhagwat@mitpune.edu.in}
}
\maketitle


\begin{abstract}
Toxic online content has become a major issue in today's world due to an exponential increase in the use of internet by people of different cultures and educational background. Differentiating hate speech and offensive language is a key challenge in automatic detection of toxic text content. In this paper, we propose an approach to automatically classify tweets on Twitter into three classes: hateful, offensive and clean. Using Twitter dataset, we perform experiments considering n-grams as features and passing their term frequency-inverse document frequency (TFIDF) values to multiple machine learning models. We perform comparative analysis of the models considering several values of \textit{n} in n-grams and TFIDF normalization methods. After tuning the model giving the best results, we achieve 95.6\% accuracy upon evaluating it on test data. We also create a module which serves as an intermediate between user and Twitter.
\end{abstract}


\section{Introduction}
\label{intro}
In the past 10 years, we have seen an exponential growth in the number of people using online forums and social networks. Every 60 seconds, there are 510,000 comments generated on Facebook \cite{1} and around 350,000 tweets generated on Twitter \cite{2}. The people interacting on these forums or social networks come from different cultures and educational backgrounds. At times, difference in opinions lead to verbal assaults. Moreover, unchecked freedom of speech over the web and the mask of anonymity that the internet provides incites people to use racists slurs or derogatory terms. This can lower the self-esteem of people, leading to mental illness and a negative impact on the society as a whole. Furthermore, toxic language can take various forms, such as cyberbullying, which was one of the major reasons behind suicide \cite{3}. This issue has shown to be increasingly important in the last decade and detecting or removing such content manually from the web is a tedious task. So there is a need of devising an automated model that is able to detect such toxic content on the web.

In order to tackle this issue, firstly we must be able to define toxic language. We broadly divide toxic language into two categories: hate speech and offensive language. Similar approach was used in the studies \cite{7} and \cite{15}. According to Wikipedia, hate speech is defined as \say{any speech that attacks a person or group on the basis of attributes such as race, religion, ethnic origin, national origin, gender, disability, sexual orientation, or gender identity.} We define offensive language as the text which uses abusive slurs or derogatory terms.

In this paper, we propose an approach to devise a machine learning model which can differentiate between these two aspects of toxic language. We choose to detect hate speech and offensive text on Twitter platform. By using publicly available Twitter datasets we train our classifier model using n-gram and term frequency-inverse document frequency (TFIDF) as features and evaluate it for metric scores. We perform comparative analysis of the results obtained using Logistic Regression, Naive Bayes and Support Vector Machines as classifier models. Our results show that Logistic Regression performs better among the three models for n-gram and TFIDF features after tuning the hyperparameters. We also make use of Twitter Application Programming Interface (API) to fetch public user tweets from Twitter for detecting tweets containing hate speech or offensive language. Additionally, we create a module which serves as an intermediate between the user and Twitter.


\section{Related Work}
\label{related_work}
Various machine learning approaches have been made in order to tackle the problem of toxic language. Majority of the approaches deal with feature extraction from the text. Lexical features such as dictionaries \cite{4} and bag-of-words \cite{5} were used in some studies. It was observed that these features fail to understand the context of the sentences. N-gram based approaches were also used which shows comparatively better results \cite{6}. 

Although lexical features perform well in detecting offensive entities, without considering the syntactical structure of the whole sentence, they fail to distinguish sentences' offensiveness which contain same words but in different orders \cite{9}. In the same study, the natural language process parser, proposed by Stanford Natural Language Processing Group, was used to capture the grammatical dependencies within a sentence.

Linguistic features such as parts-of-speech has also been used in hate speech detection problem, as shown in \cite{8}; these approaches consist in detecting the category of the word, for instance, personal pronoun (PRP), Verb non-3rd person singular present form (VBP), Adjectives (JJ), Determiners (DT), Verb base forms (VB).

There have been several studies on sentiment-based methods to detect abusive language published in the last few years. One example is the work \cite{9} which applies sentiment analysis to detect bullying in tweets and use Latent Dirichlet Allocation (LDA) topic models \cite{10} to identify relevant topics in these texts. Also studies have been conducted for Detection of harassment on Web 2.0 \cite{11}

More recently, distributed word representations, also referred to as word embeddings, have been proposed for a similar purposes \cite{12}. Deep learning techniques are recently being used in text classification and sentiment analysis using paragraph2vec approach \cite{13}. Convolutional Neural Network (CNN) based classification, which refers to the generation of a CNN for text classification, is being used as seen in \cite{14}, where they experimented with a system for Twitter hate-speech text classification based on a deep-learning, CNN model.

\section{Proposed Approach}
\label{proposed_approach}
The review on the related work done in this field shows that the models trained after extracting N-gram features from text give better results \cite{6}. Also, the TFIDF approach on the bag-of-words features also show promising results \cite{5}. Based on the review of features and the prominent classifiers used for text classification in the past work, we decided to extract n-grams from the text and weight them according to their TFIDF values. We feed these features to a machine learning algorithm to perform classification. Given the set of tweets, the aim of this work is to classify them into three categories: hateful, offensive and clean.

\subsection{Data}
The dataset that we have generated is a combination of three different datasets. The first dataset is publicly available on Crowdflower\footnote{\url{https://data.world/crowdflower/hate-speech-identification}}, which was used in \cite{7} and \cite{15}. This dataset contains tweets that have been manually classified into one of the following classes: ``Hateful'', ``Offensive'' and ``Clean''. The second dataset is also publicly available on Crowdflower\footnote{\url{https://data.world/ml-research/automated-hate-speech-detection-data}}, which consists the tweets with same classes as described previously. The third dataset is published on Github\footnote{\url{https://github.com/ZeerakW/hatespeech}} and used in the work \cite{7} and \cite{16}. It consists of two columns: tweet-ID and class. In this dataset, tweets corresponding to the tweet-ID are classified into one of the following three classes: ``Sexism'', ``Racism'' and ``Neither''.

\subsection{Data Preprocessing}
In the data preprocessing stage, we combine the three datasets used for this work. The tasks involves removal of unnecessary columns from the datasets and enumerating the classes. For the third dataset, we retrieve the tweets corresponding to the tweet-ID present in the dataset. We use Twitter API for this purpose. The classes ``Sexism'' and ``Racism'' in this dataset are both considered as hate speech according to the definition.

We convert the tweets to lowercase and remove the following unnecessary contents from the tweets:
\begin{itemize}
\item Space Pattern
\item URLs
\item Twitter Mentions
\item Retweet Symbols
\item Stopwords
\end{itemize}
We use the Porter Stemmer algorithm to reduce the inflectional forms of the words.

After combining the dataset in proper format, we randomly shuffle and split the dataset into two parts: train dataset containing 70\% of the samples and test dataset containing 30\% of the samples.

\subsection{Feature Extraction}
We extract the n-gram features from the tweets and weight them according to their TFIDF values. The goal of using TFIDF is to reduce the effect of less informative tokens that appear very frequently in the data corpus. Experiments are performed on values of $n$ ranging from one to three. Thus, we consider unigram, bigram and trigram features. The formula that is used to compute the TFIDF of term $t$ present in document $d$ is: \[tfidf(d,t) = tf(t)\;  * \;idf(d,t)\] Also, both L1 and L2 (Euclidean) normalization of TFIDF is considered while performing experiments. L1 normalization is defined as: \[v_{norm} = \frac{v}{|v_1|+|v_2|+...+|v_n|}\] where $n$ in the total number of documents. Similarly, L2 normalization is defined as: \[v_{norm} = \frac{v}{\sqrt{v_1^2+v_2^2+...+v_n^2}}\] We feed these features to machine learning models.

\subsection{Model}
We consider three prominent machine learning algorithms used for text classification: Logistic Regression, Naive Bayes and Support Vector Machines. We train each model on training dataset by performing grid search for all the combinations of feature parameters and perform 10-fold cross-validation. The performance of each algorithm is analyzed based on the average score of the cross-validation for each combination of feature parameters. The performance of these three algorithms is compared.

Further, the hyperparameters of two algorithms giving best results are tuned for their respective feature parameters, which gives the best result. Again, 10-fold cross validation is performed to measure the results for each combination of hyperparameters for that model. The model giving the highest cross-validation accuracy is evaluated against the test data. We have used scikit-learn in Python for the purpose of implementation.

\section{Results}
\label{results}
\begin{figure*}
	\centering
	\includegraphics[width=\textwidth, height=7cm]{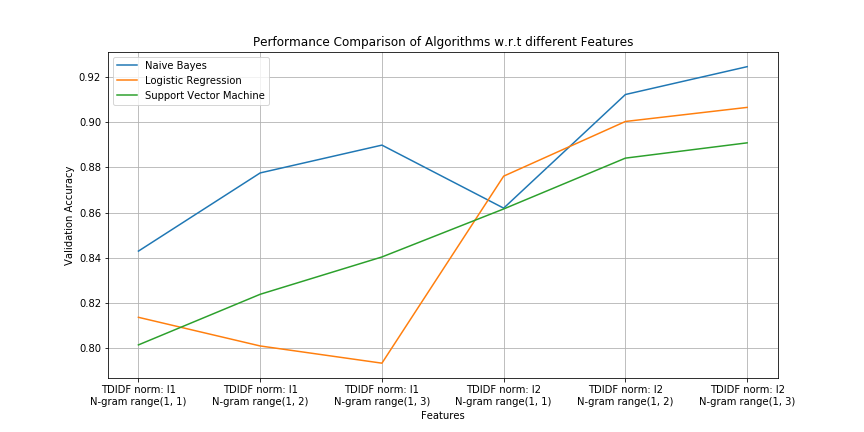}
	\caption{The figure shows comparative analysis of Naive Bayes, SVM and Logistic Regression on various sets of feature parameters}
	\label{fig:performance_comparison}
\end{figure*}

The results of the comparative analysis of Logistic Regression (LR), Naive Bayes (NB) and Support Vector Machines (SVM) for various combinations of feature parameters is shown in Fig. \ref{fig:performance_comparison} and TABLE \ref{table:performance_comparison}.

\begin{table}
	\centering
    \caption{Comparison of three models for different combinations of feature parameters}
    \begin{tabular}{ |c|c|c|c| } 
    \hline
    \multirow{3}{*}{N-gram Range + TFIDF Norm} & \multicolumn{3}{c|}{Accuracy} \\
    \cline{2-4}
    & NB & LR & SVM \\
   	\hline
    (1,1) + L1 & 0.842 & 0.816 & 0.802 \\
    \hline
    (1,2) + L1 & 0.878 & 0.801 & 0.823 \\
    \hline
    (1,3) + L1 & 0.890 & 0.794 & 0.841 \\
    \hline
    (1,1) + L2 & 0.862 & 0.878 & 0.862 \\
    \hline
    (1,2) + L2 & 0.913 & 0.901 & 0.884 \\
    \hline
    (1,3) + L2 & \textbf{0.926} & \textbf{0.918} & \textbf{0.901} \\
    \hline
    \end{tabular}
    \label{table:performance_comparison}
\end{table}

Fig. \ref{fig:performance_comparison} shows that all the three algorithms perform significantly better for the L2 normalization of TFIDF. However, SVM performs poorly as compared to Naive Bayes and Logistic Regression for L2 normalization. TABLE \ref{table:performance_comparison} shows that the best result for Naive Bayes, 92.6\%, is obtained using n-gram range up to three and TFIDF normalization L2. Similarly, Logistic Regression performs better for the same set of feature parameters achieving 91.3\% accuracy. Since both of these values are comparable, we tune both Naive Bayes and Logistic Regression, for the n-gram range up to three and TFIDF normalization L2.

TABLE \ref{table:naive_bayes_tuned} shows the results after tuning the Naive Bayes algorithm. We have considered the smoothing prior $\alpha$ for tuning. $\alpha \geq 0$ considers the features which are not present in the training set and in turn prevents zero probabilities. Technically, $\alpha = 1$ is called Laplace smoothing and $\alpha < 1$ is called Lidstone smoothing. Naive Bayes performs better for the $\alpha$ value 0.1 giving 93.4\% accuracy.

\begin{table}
	\centering
    \caption{Results after tuning Naive Bayes w.r.t smoothing prior $\alpha$ for the features: n-gram range 1-3 and TFIDF normalization L2}
    \begin{tabular}{ |c|c| } 
    \hline
    Alpha ($\alpha$) & Accuracy \\
   	\hline
    0.01 & 0.931 \\
    \hline
    0.1 & \textbf{0.934} \\
    \hline
    1 & 0.925 \\
    \hline
    10 & 0.877 \\
    \hline
    \end{tabular}
    \label{table:naive_bayes_tuned}
\end{table}

TABLE \ref{table:logistic_regression_tuned} shows the performance after tuning the Logistic Regression algorithm. Here, we have considered the regularization parameter $C$ and the optimization algorithms (solvers) -- liblinear, newton-cg and saga -- for performance tuning. The model with settings $C=100$ and solver liblinear gives the best accuracy 95.1\%.

\begin{table}
	\centering
    \caption{Results after tuning Logistic Regression w.r.t regularization parameter $C$ and various optimization algorithms (solvers) for the features: n-gram range 1-3 and TFIDF normalization L2}
    \begin{tabular}{ |c|c| } 
    \hline
    Regularization $C$ + Solver & Accuracy \\
   	\hline
    10 + liblinear & 0.949 \\
    \hline
    10 + newton-cg & 0.948 \\
    \hline
    10 + saga & 0.948 \\
    \hline
    100 + liblinear & \textbf{0.951} \\
    \hline
    100 + newton-cg & 0.950 \\
    \hline
    100 + saga & 0.950 \\
    \hline
    \end{tabular}
    \label{table:logistic_regression_tuned}
\end{table}

Comparing the best accuracy for Naive Bayes and Logistic Regression, we conclude that Logistic Regression performs better. Therefore, we evaluate Logistic Regression on test data with the settings: n-gram range 1-3, TFIDF normalization L2, $C=100$ and optimization algorithm liblinear. The classification scores are shown in TABLE \ref{table:classification_scores}.

\begin{table}
	\centering
    \caption{Classification scores obtained after evaluating the final Logistic Regression model on test data.}
    \begin{tabular}{ |c|c|c|c| } 
    \hline
    & Precision & Recall & F-score \\
   	\hline
    Hateful & 0.94 & 0.96 & 0.95 \\
    \hline
    Offensive & 0.96 & 0.93 & 0.94 \\
    \hline
    Clean & 0.96 & 0.98 & 0.97 \\
    \hline
    Average & 0.96 & 0.96 & 0.96 \\
    \hline
    \end{tabular}
    \label{table:classification_scores}
\end{table}

It is observed that the recall for offensive text is relatively low, 0.93. This means that 7\% of the tweets that are actually offensive have been misclassified by the model. Also, the precision for the hateful class is 0.94, which signifies that 6\% of the tweets that are either clean or offensive have been classified as hateful. On the other hand, the recall for clean class is 0.98, which is significantly better.

\begin{table}
	\centering
    \caption{Confusion Matrix for the evaluated test data on the final Logistic Regression model}
    \begin{tabular}{ |c|c|c|c| } 
    \hline
    \multirow{2}{*}{Class} & \multicolumn{3}{c|}{Classified as} \\
   	\cline{2-4}
    & Hateful & Offensive & Clean \\
    \hline
    Hateful & 0.965 & 0.021 & 0.014 \\
    \hline
    Offensive & 0.048 & 0.926 & 0.026 \\
    \hline
    Clean & 0.010 & 0.013 & 0.977 \\
    \hline
    \end{tabular}
    \label{table:confusion_matrix}
\end{table}

In addition to the classification scores, we also computed the confusion matrix for the test results which is shown in TABLE \ref{table:confusion_matrix}. The key point to notice here is that 4.8\% of the tweets that are offensive have been classified as hateful. Improvements can be done in this area to further increase the scores of the model. The final testing accuracy of the model is obtained to be 95.6\%.

\section{Interfacing with Twitter}
\label{interface}
Our final model is configured to interface with Twitter through the use of Twitter API particularly to collect data tweets via Twitter REST API. In python, the library Tweepy helps add this functionality with simplicity. Twitter APIs, besides basic information such as the tweet text and the author of the tweet, returns data structure contains additional information which can be used to provide further analysis. For each maximum 140 character tweet, API returns a JSON document containing several items of metadata presented as key and value pairs, out of which \textit{id} and \textit{text} are most important for the sake of this study.

\begin{figure}
  \centering
  \includegraphics[width=0.5\textwidth]{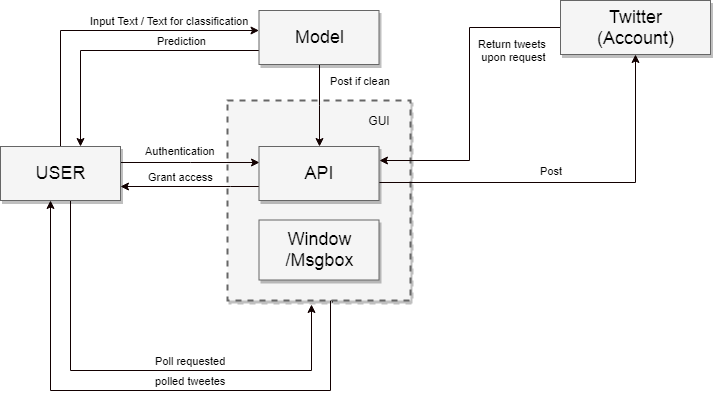}
  \caption{Architecture of the system interfacing with Twitter through Twitter API}
  \label{fig:architecture}
\end{figure}

We also create an application which acts as a module between the user and Twitter. The architecture of the application is shown in Fig. \ref{fig:architecture}. Through our module, we are able to filter out hateful and offensive tweets being posted by an individual as well as classify the tweets posted on the user home timeline, with the only limitation being twitter read request rate limiter of 15 minutes.

\section{Conclusion}
\label{conclusion}
In this paper, we proposed a solution to the detection of hate speech and offensive language on Twitter through machine learning using n-gram features weighted with TFIDF values. We performed comparative analysis of Logistic Regression, Naive Bayes and Support Vector Machines on various sets of feature values and model hyperparameters. The results showed that Logistic Regression performs better with the optimal n-gram range 1 to 3 for the L2 normalization of TFIDF. Upon evaluating the model on test data, we achieved 95.6\% accuracy. It was seen that 4.8\% of the offensive tweets were misclassified as hateful. This problem can be solved by obtaining more examples of offensive language which does not contain hateful words. The results can be further improved by increasing the recall for the offensive class and precision for the hateful class. Also, it was seen that the model does not account for negative words present in a sentence. Improvements can be done in this area by incorporating linguistic features.


\end{document}